\documentclass[english]{article}
\usepackage[T1]{fontenc}
\usepackage[latin9]{inputenc}
\usepackage{color}
\usepackage{babel}
\usepackage{amsmath}
\usepackage{graphicx}
\usepackage[unicode=true,pdfusetitle,
 bookmarks=true,bookmarksnumbered=false,bookmarksopen=false,
 breaklinks=false,pdfborder={0 0 1},backref=false,colorlinks=true]
 {hyperref}

\makeatletter

\newcommand*\LyXZeroWidthSpace{\hspace{0pt}}
\providecommand{\tabularnewline}{\\}

\@ifundefined{date}{}{\date{}}
\usepackage{microtype}
\usepackage{graphicx}
\usepackage{booktabs} 

\usepackage{hyperref}

\usepackage[accepted]{icml2020}

\icmltitlerunning{DeepCoDA: personalized interpretability for compositional health data}

\makeatother

\begin{document}
\twocolumn[ 
\icmltitle{DeepCoDA: personalized interpretability for compositional health data}
\icmlsetsymbol{equal}{*}
\begin{icmlauthorlist} 
\icmlauthor{Thomas P. Quinn}{equal,to} 
\icmlauthor{Dang Nguyen}{equal,to} 
\icmlauthor{Santu Rana}{to} 
\icmlauthor{Sunil Gupta}{to} 
\icmlauthor{Svetha Venkatesh}{to}
\end{icmlauthorlist}
\icmlaffiliation{to}{Applied Artificial Intelligence Institute (A\textsuperscript{2}I\textsuperscript{2}), Deakin University, Geelong, Australia}
\icmlcorrespondingauthor{Thomas P. Quinn}{contacttomquinn@gmail.com} 
\icmlkeywords{Machine Learning, ICML}
\vskip 0.3in 
]

\printAffiliationsAndNotice{\icmlEqualContribution}
\begin{abstract}
Interpretability allows the domain-expert to directly evaluate the
model's relevance and reliability, a practice that offers assurance
and builds trust. In the healthcare setting, interpretable models
should implicate relevant biological mechanisms independent of technical
factors like data pre-processing. We define \emph{personalized interpretability}
as a measure of sample-specific feature attribution, and view it as
a minimum requirement for a precision health model to justify its
conclusions. Some health data, especially those generated by high-throughput
sequencing experiments, have nuances that compromise precision health
models and their interpretation. These data are \emph{compositional},
meaning that each feature is conditionally dependent on all other
features. We propose the \textbf{Deep Co}mpositional \textbf{D}ata
\textbf{A}nalysis \textbf{(DeepCoDA)} framework to extend precision
health modelling to high-dimensional compositional data, and to provide
personalized interpretability through patient-specific weights. Our
architecture maintains state-of-the-art performance across 25 real-world
data sets, all while producing interpretations that are both personalized
and fully coherent for compositional data.
\end{abstract}

\section{Introduction\label{sec:Introduction}}

\inputencoding{latin9}Interpretability is pivotal for the adoption
and use of predictive deep models in healthcare. As one physician
noted, ``Without {[}English-language{]} explanations, it is obviously
unreasonable for the physician to rely on such programs; ultimately,
a program, like any consultant, must justify its conclusions to the
physician responsible for the patient's care'' \cite{schwartz_artificial_1987}.
Interpretability allows the domain-expert to directly evaluate the
model's relevance and reliability, a practice that offers assurance
and builds trust. In the healthcare setting, interpretable models
should implicate relevant biological mechanisms independent of technical
factors like data pre-processing. Precision health research aims to
target disease prevention and health promotion to individual patients
\cite{ashley_towards_2016}. Although this field has benefited tremendously
from the availability of clinical data, its mission requires new models
that offer interpretability at the level of the individual patient.
We define \emph{personalized interpretability} as a measure of sample-specific
feature attribution, and view it as a minimum requirement for a precision
health model to justify its conclusions.

The \textit{attention mechanism} offers one approach to personalized
interpretability \cite{bahdanau_neural_2016}, and has been used recently
to study cancer heterogeneity \cite{beykikhoshk_deeptriage_2020}.
The \emph{self-explaining neural network} (SENN), may likewise enable
precision health through its formulation $y=\theta(\mathbf{x})^{\top}\mathbf{x}$
\cite{alvarez-melis_towards_2018}. This equation is well-suited to
personalized interpretability for 3 reasons: (a) the input features
$x_{j}$ are clearly anchored to the observed empirical measurements;
(b) each parameter $\theta(\mathbf{x})_{j}$ estimates the quantitative
contribution of its corresponding feature $x_{j}$ to the predicted
value; and (c) the aggregation of feature specific terms $\theta(\mathbf{x})_{j}x_{j}$
is additive, allowing for a feature-by-feature interpretation of impact.
However, self-explanation has not yet been tailored to high-throughput
health biomarker data, which exist not as absolute measurements but
rather multivariate compositions.

Much of the health data generated by high-throughput sequencing experiments
have nuances that compromise precision health models and their interpretation.
These data are compositional, meaning that they arise from an inexhaustive
sampling procedure in which each feature is conditionally dependent
on all other features \cite{gloor_microbiome_2017,quinn_understanding_2018,calle_statistical_2019}.
Compositional data can be defined as the proportions of $\textbf{x}^{*}$
\cite{aitchison_statistical_1986}
\begin{equation}
\textbf{x}=\frac{\left[x_{1}^{*},...,x_{D}^{*}\right]}{\sum_{j}^{D}x_{j}^{*}}
\end{equation}
where $\textbf{x}$ is a composition with $D$ parts. By considering
this equation carefully, one can gain an intuition for why compositional
data are so difficult to analyze. For example, consider the 3-part
vector $\textbf{x}^{*}=[a^{*},b^{*},c^{*}]$ of \textit{absolute}
abundances and its corresponding composition $\textbf{x}=[a,b,c]$
of \textit{relative} abundances. When $c^{*}$ increases, $c$ will
also increase (as expected). However, $a$ and $b$ must also decrease
because the sum of $\textbf{x}$ is fixed! As such, the value of any
one part $x_{j}$ depends on all other parts. For compositional data,
common measures of association \cite{pearson_mathematical_1896,lovell_proportionality:_2015},
distance \cite{aitchison_logratio_2000}, and feature attribution
\cite{boogaart_descriptive_2013} can be misleading. Likewise, both
supervised learning \cite{rivera-pinto_balances:_2018,tolosana_delgado_machine_2019}
and unsupervised learning \cite{avalos_representation_2018,martin-fernandez_advances_2019}
require an innovative approach.

In practice, analysts use normalization in an attempt to make the
data absolute, but these rely on untestable assumptions (e.g., that
the majority of features remain unchanged \cite{robinson_scaling_2010,anders_differential_2010}).
Two manuscripts, both studying bacteria-metabolite associations,
avoided normalization by having the hidden layers compute on log-ratio
transformed data, either implicitly (through an inverse transform
of the output layer \cite{morton_learning_2019}) or explicitly (through
a transform of the input layer \cite{le_deep_2019}). The latter model
sought interpretability through heavy regularization and a non-negative
weights constraint, but its interpretation would still depend on a
normalizing assumption even if its performance does not (c.f., \cite{erb_how_2016}).
\emph{We are unaware of any neural network architecture designed specifically
for the personalized interpretation of compositional data.}

We propose a normalization-free neural network architecture called
\textbf{DeepCoDA} to provide personalized interpretability for compositional
data. We overcome 3 key challenges through the use of an end-to-end
neural network:
\begin{itemize}
\item \textbf{A model should select the best log-ratio transformation automatically.}
It is necessary to transform the feature space, but there are many
ways to do this. We propose a \emph{log-bottleneck module} to learn
useful log-contrasts from the training data. It works by passing log-transformed
data through a hidden layer with a single node such that the layer
weights sum to 0. The node thus becomes a simple log-contrast, and
$B$ modules can be stacked in parallel to get $B$ log-contrasts. 
\item \textbf{A model should have linear interpretability. } The use of
non-linear transformations will often improve the predictive performance
of a model, but this can come at the expense of interpretability.
We use a \emph{self-explanation module} to introduce linear interpretability,
which has the form $y=\theta(\mathbf{x})^{\textrm{\ensuremath{\top}}}\mathbf{x}$
\cite{alvarez-melis_towards_2018}. Each output is predicted by a
linear combination of the input, just like a linear regression.
\item \textbf{A model should have personalized interpretability.} Many models
can estimate feature importance for the whole population, but precision
health requires an estimation of importance for the individual patient.
This is also achieved by the \emph{self-explanation module} (mentioned
above), which computes patient-specific weights as a function of the
input.
\end{itemize}
Biological scientists routinely study the gut microbiome as a biomarker
for disease prediction and surveillance (e.g., inflammatory bowel
disease and cancer \cite{duvallet_meta-analysis_2017}). Microbiome
data are compositional, and thus inherit the data nuances described
above. We validate our \textbf{DeepCoDA} framework on 25 separate
microbiome and similar data sets. Using a cohort of 13 data sets,
we find a set of hyper-parameters that work generally well, and verify
them on 12 unseen data sets. This two-step procedure allows us to
recommend general hyper-parameters for real-world microbiome data.

Our framework solves an open problem in biology: how to extend personalized
interpretability to compositional health data, including RNA-Seq,
metagenomics, and metabolomics. The novelty lies in using neural networks
to learn predictive log-contrasts, and coupling them with self-explanation
to provide personalized interpretability through simple algebraic
expressions. This combined architecture results in interpretations
that are both personalized and fully coherent for compositional data,
all while maintaining state-of-the-art performance.
\vspace{3mm}

\section{Related Background\label{sec:Related-Work}}

\inputencoding{latin9}\textbf{Compositional data analysis:} The field
of compositional data analysis (CoDA) emerged in the 1980s \cite{aitchison_statistical_1986}.
Most often, CoDA uses an internal reference to transform the raw variables
into a set of log-ratios for which the denominator is the reference.
Log-ratio models are implicitly normalization-free because the normalization
factor cancels by the ratio. Popular references include the per-sample
geometric mean (\emph{centered log-ratio transform}), a single feature
of importance (\emph{additive log-ratio transform}), or all features
(\emph{pairwise log-ratio transform}) \cite{aitchison_statistical_1986}.
We include the centered log-ratio as part of the baseline used to
benchmark our model.

\textbf{Log-contrast models:} \citet{aitchison_log_1984} proposed
log-contrast models for the analysis of compositional data. According
to the authors, log-contrasts are useful when an analyst wants to
study the change of some parts of a composition while holding fixed
some other parts. Like log-ratio models, log-contrast models are normalization-free
because the normalization factor cancels. A subset of log-contrasts,
called \emph{balances}, have gained some popularity in microbiome
research \cite{morton_balance_2017,washburne_phylogenetic_2017,silverman_phylogenetic_2017}.
Recent work has proposed data-driven heuristics, including \emph{forward-selection}
\cite{rivera-pinto_balances:_2018} and \emph{cluster analysis} \cite{quinn_interpretable_2020},
to identify predictive balances. These approaches take a statistical
learning perspective that may not generalize to non-linear multivariable
regression. We include balances as part of the baseline used to benchmark
our model.

\textbf{Parallel blocks:} \citet{tsang_neural_2018} proposed the
Neural Interaction Transparency (NIT) architecture to discover statistical
interactions among variables. This architecture uses multiple multi-layer
perceptrons stacked as parallel \textit{blocks}, where incoming connections
define \textit{feature sets}. Our proposed network also uses the idea
that parallel blocks can define interpretable feature sets. For NIT,
the authors interpret each feature set as a statistical interaction.
They learn each feature set in parallel, and the output is predicted
by an additive combination of the feature sets. For \textbf{DeepCoDA},
we interpret each feature set as a single log-contrast. We learn these
log-contrasts in parallel, and the output is also predicted by an
additive combination of the log-contrasts.

\textbf{Attention:} \citet{bahdanau_neural_2016} used an attention
mechanism for neural machine translation. Within a recurrent neural
network, they compute a context vector for each time state $i$ as
the weighted sum of the encoded input. The weights are determined
by an attention vector $\alpha$ as a function of the latent variables
themselves. One could interpret the attention vector weights as a
measure of feature importance, and thus view attention as an estimate
of sample-specific importance. As such, attention is highly relevant
to precision health, which aims to tailor medical care to the individual
patient. \citet{beykikhoshk_deeptriage_2020} used attention to classify
gene expression signatures, and analyzed the attention vectors directly
to study disease heterogeneity.

\textbf{Self-explaining neural networks:} \citet{alvarez-melis_towards_2018}
proposed the self-explaining neural network as a highly interpretable
architecture that acts like a simple linear model locally, but not
globally. They describe a linear model whose coefficients depend on
the input itself, having the form $f(\mathbf{x})=\theta(\mathbf{x})^{\textrm{\ensuremath{\top}}}\mathbf{x}$.
When $\theta$ is a neural network, this function can learn complex
non-linear functions while still having the interpretability of a
linear model. The authors extend this function beyond input features
to latent features, and propose a regularization penalty to force
the model to act locally linear. In our work, we use the form $f(\boldsymbol{\textbf{z}})=\theta(\boldsymbol{\textbf{z}})^{\textrm{T}}\boldsymbol{\textbf{z}}$
to learn sample-specific weights for each log-contrast in \textbf{$\boldsymbol{\mathbf{z}}$}.
This is similar to computing an attention vector, but we use self-explanation
because it is algebraically and conceptually simpler. 
\vspace{3mm}

\section{The Proposed Framework\label{sec:Framework}}

\subsection{Network architecture}

\inputencoding{latin9}\begin{figure}
\centering{}\includegraphics[scale=0.42]{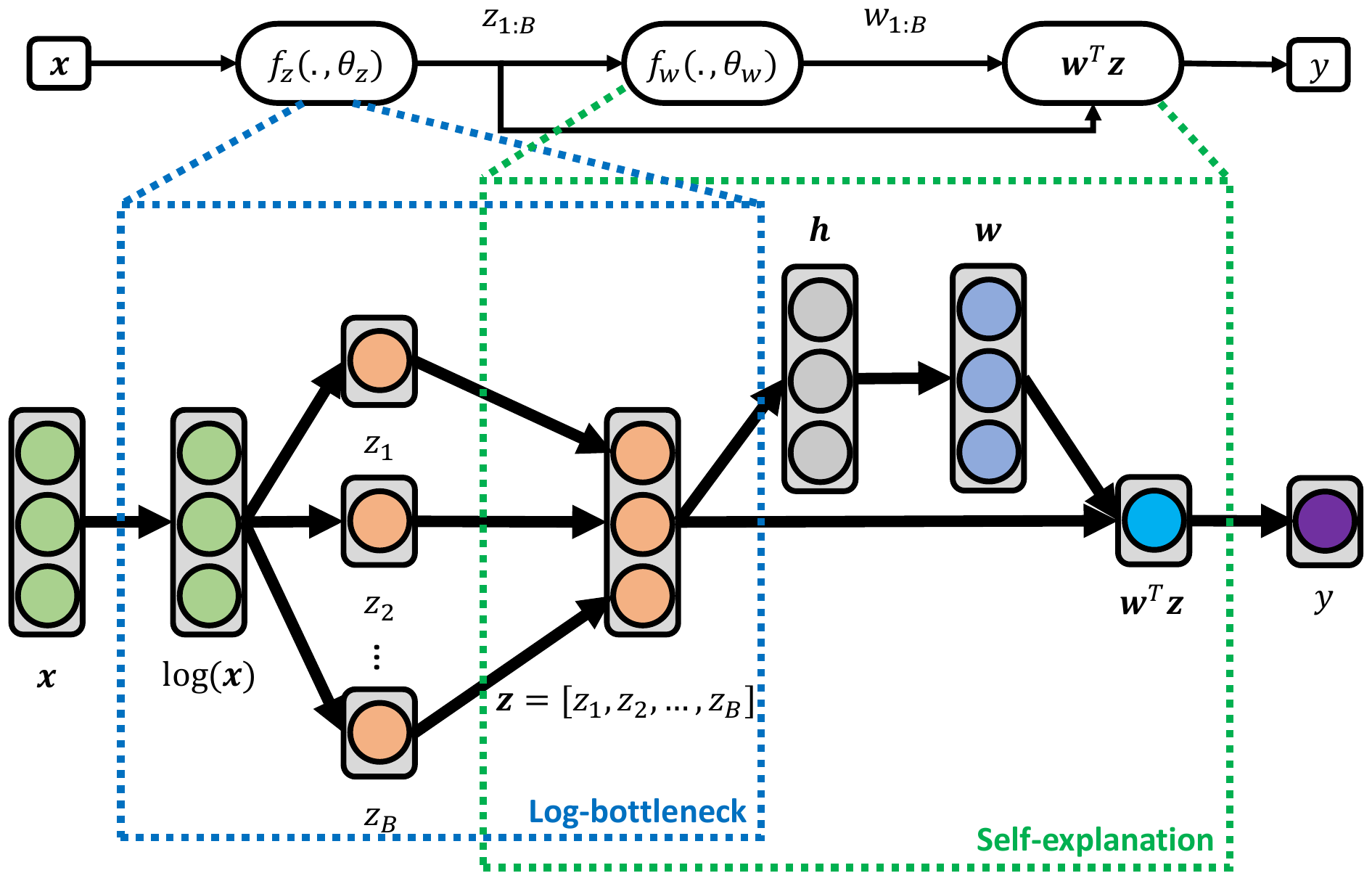}\caption{The network architecture has two distinct modules. The first is the
\emph{log-bottleneck module} which passes the log-transformed data
through a bottleneck layer of a single hidden node such that the layer
weights sum to 0. This hidden node therefore describes a single log-contrast,
and $B$ modules can be stacked in parallel to get $B$ log-contrasts.
The second is the \emph{self-explanation module} which introduces
both linear interpretability and personalized interpretability.}
\label{fig:overview}
\end{figure}
Figure~\ref{fig:overview} provides a visual overview of our proposed
neural network architecture for compositional data analysis. It contains
2 distinct modules.

\subsubsection{The log-bottleneck module}

\begin{align}
\textbf{z}_{i}=f_{z}(\textbf{x}_{i};\theta_{z})= & [z_{i1},...,z_{iB}]
\end{align}

The first module is the \emph{log-bottleneck.} The function $f_{z}$
takes a composition of $d=1...D$ features as input and returns $b=1...B$
log-contrasts, parameterized by $\theta_{z}$. This module is described
in detail in Section \ref{module1}.

\subsubsection{The self-explanation module}

\begin{align}
\textbf{w}_{i}= & f_{w}(\textbf{z}_{i};\theta_{w})=[w_{i1},...,w_{iB}]\\
y_{i}= & f_{a}(\textbf{w}_{i},\textbf{z}_{i};\theta_{a})=\textbf{w}_{i}^{\textrm{T}}\textbf{z}_{i}
\end{align}

The second module is the \emph{self-explanation}. The function $f_{w}$
takes a vector of $b=1...B$ log-contrasts as input and returns $b=1...B$
weights, parameterized by $\theta_{w}$. We define $f_{w}$ as a neural
network with a single hidden layer of 16 nodes and ReLu activation.
The function $f_{a}$ is a simple dot product between the input-specific
weight vector $\textbf{w}_{i}$ and the log-contrasts $\textbf{z}_{i}$,
so $\theta_{a}$ is a null set. Thus $y_{i}=\textbf{w}_{i}^{\textrm{T}}\textbf{z}_{i}$,
which one could extend to multivariable output. This module is described
in detail in Section \ref{module2}.

\subsection{Log-contrasts and sub-compositional coherence\label{module1}}

The parts of a composition are fundamentally intertwined. It is impossible
to make inferences about the absolute abundance of one variable $x_{j}^{*}$
based on the component $x_{j}$. This is because the magnitude of
$x_{j}$ depends on all $x_{k\neq j}$. In order to make inferences
from relative data that are also true for the absolute data (and thus
true in an absolute sense) one \textit{must} use a reference.

By using a ratio of components, the denominator serves as a reference.
The ratio $a/b$ describes the abundance of $a$ relative to $b$
(or $a$ conditional upon $b$ \cite{aitchison_log_1984}), while
the logarithm is used to center the ratio so that $\log(a/b)=0$ when
$a=b$. Note here that this log-contrast does not depend on the value
of $c$ (or $c^{*}$). Log-ratios and log-contrasts are \emph{sub-compositionally
coherent}: the results do not change when analyzing a subset of the
composition \cite{boogaart_fundamental_2013}. Any inference made
on a ratio of relative data will agree with an inference made on a
ratio of absolute data.

A l\emph{og-contrast model} \cite{aitchison_log_1984} is simply a
linear model of the log-transformed data:

\begin{equation}
y_{i}=\beta_{0}+\sum_{d=1}^{D}\beta_{d}\log x_{id}
\end{equation}
By forcing that $\sum{}_{d=1}^{D}\beta_{d}=0$, the log-contrast becomes
more interpretable because the coefficients of the numerator and denominator
both have equal weight.

Our network contains a \emph{log-bottleneck module} capable of learning
useful log-contrasts. This module passes a log-transform of the input
data through a bottleneck layer of a single hidden unit.

\begin{align}
z_{ib}= & \beta_{0}^{(b)}+\beta_{1}^{(b)}\log x_{i1}+...+\beta_{D}^{(b)}\log x_{iD}\\
= & \beta_{0}^{(b)}+\log\Bigg(\prod_{d=1}^{D}(x_{id})^{\beta_{d}^{(b)}}\Bigg)
\end{align}
for sample $i$ and bottleneck $b$, where $z_{ib}$ is a single hidden
unit. By adding a loss that constrains $\sum_{d=1}^{D}\beta_{d}^{(b)}=0$,
then $z_{ib}$ approximates a log-contrast where the neural network
weights refer to the powers (i.e., because $a\log(b)=\log(b^{a})$).
One could repeat $b={1...B}$ bottlenecks in parallel, depending on
the number of log-contrasts required. A regularization penalty determines
how many variables comprise each log-contrast, where a larger penalty
results in fewer parts. Note, each log-bottleneck resembles the \emph{coda-lasso
}model from \cite{susin_variable_2020}.

\subsection{Linearity and self-explanation\label{module2}}

One advantage of a neural network over a simple linear model is that
it can learn a complex non-linear mapping between multivariable input
and multivariable output. However, linear models are often preferred
in the applied setting because the weights can be interpreted directly
as a kind of feature importance. A \emph{(generalized) linear model}
(GLM) \cite{nelder_generalized_1972} has the form

\begin{equation}
y_{i}=\Phi\big(w_{0}+w_{1}\textbf{x}_{i1}+...+w_{D}\textbf{x}_{iD}\big)
\end{equation}
where $\textbf{w}_{d}$ is the weight for feature $d$ and $\Phi$
is a monotonic transform. For example, $\Phi$ may be a logistic function
when $y$ is binary. By transforming the entire output of the linear
model, the ``link'' function $\Phi$ expands the solution space,
thus ``generalizing'' the model. A \emph{generalized additive model}
(GAM) \cite{hastie_generalized_1986} expands the solution space further,
allowing for a more complex mapping between the input and output.
A GAM has the form

\begin{equation}
y_{i}=\Phi\big(w_{0}+f_{1}(\textbf{x}_{i1})+...+f_{D}(\textbf{x}_{iD})\big)
\end{equation}
where $f_{d}$ is a non-linear function of a single input variable
$x_{id}$. A GAM is potentially more powerful than a GLM, yet still
interpretable because each weight describes the importance of a single
feature (or a function thereof). However, this model can only solve
equations in which the predictor variables make an independent contribution
to the output. As written, a GAM cannot model the multiplicative interactions
between two or more input variables, e.g., where $y=x_{1}+x_{2}+x_{1}*x_{2}$.

On the other hand, a fully-connected neural network can learn the
interactions between variables without having to specify them explicitly.
However, neural networks are hard to interpret because they involve
a series of non-linearly transformed matrix products, making it difficult
to know how or why a feature contributed to the prediction.

To overcome the disadvantages of GLM, GAM, and neural networks, we
introduce a \emph{self-explanation module} which adapts the key idea
behind the self-explaining neural network (SENN) \cite{alvarez-melis_towards_2018}.
Our network uses self-explanation to introduce non-linearity into
the model while also providing personalized interpretability. It has
the form

\begin{align}
y_{i}= & \theta(\textbf{x}_{i})^{\textrm{T}}\cdot\textbf{x}_{i}\\
= & w_{i1}x_{i1}+...+w_{iD}x_{iD}
\end{align}
for sample $i$, where $\theta$ is a neural network. Here, $x_{id}$
is the abundance of feature $d$ for sample $i$ while $w_{id}$ is
the weight of feature $d$ for sample $i$. Note that for GLM and
GAM, each weight $w_{d}$ is defined for an entire population. For
self-explanation, each weight $w_{id}$ is defined for an individual
sample. As such, we can interpret $w_{id}$ as a kind of personalized
importance score.

We apply self-explanation to the log-contrasts $z_{ib}$:

\begin{align}
y_{i}= & \theta(\textbf{z}_{i})^{\textrm{T}}\cdot\textbf{z}_{i}\\
= & w_{i1}z_{i1}+...+w_{iB}z_{iB}
\end{align}

\citet{alvarez-melis_towards_2018} proposed a regularization penalty
that forces the self-explanation to act locally linear. We did not
include this penalty, and instead throttled the complexity of $\theta$.
This design choice reduces the number of parameters and hyper-parameters,
making it easier to learn with so few samples.

\subsection{Loss}

The network is fully differentiable and trained end-to-end. The loss
has 3 parts: the mean-squared error, the log-contrast constraint,
and the L1-norm regularization penalty.

\begin{align}
\mathcal{L}_{T}= & \sum_{i=1}^{N}(\hat{y}_{i}-y_{i})^{2}+\sum_{b=1}^{B}\lambda_{c}(\sum_{d=1}^{D}\beta_{db})^{2}+\sum_{b=1}^{B}\lambda_{s}(\sum_{d=1}^{D}|\beta_{db}|)
\end{align}
where the weights $\beta_{db}$ come from the log-bottleneck $f_{z}$.
Here, $\lambda_{c}$ is very large to force the weights of the log-contrast
to sum to 0, while $\lambda_{s}$ makes it so that each log-contrast
contains fewer parts. The number of log-bottlenecks $B$ and the complexity
of each log-contrast $\lambda_{s}$ are two important hyper-parameters
for this model. We set $\lambda_c = 1$.
\vspace{3mm}

\section{Experiments\label{sec:Experiments}}

\subsection{Data and baselines}

\subsubsection{Synthetic data}

\inputencoding{latin9}We simulate 2 synthetic data sets, both containing
1000 samples belonging to 2 classes (``case'' vs. ``control'').
For the simulated data, we generate the absolute abundances, then
divide each sample by its total sum to obtain the relative abundances.
For real data, we do not know the absolute abundances, so the simulated
data grants us a unique opportunity to compare absolute and relative
data analyses.

The first simulated data set is a toy example. It contains 4 variables
that represent different bacteria within the gut of a patient, such
that the over-proliferation of bacteria \{2, 3, 4\} cause a disease.
Our task is to predict whether the simulated patient has a sick or
healthy gut.

The second is based on a real biological example in which a mutation
of the c-Myc gene causes a cell to massively increase the production
of 90\% of its gene transcripts, while keeping 10\% the same \cite{loven_revisiting_2012}.
Our task is to predict whether the simulated cell is c-Myc positive
or c-Myc negative. Although the original data contain 1000s of features,
our simulated version has only 10 features to make visualization easier.

\subsubsection{Real data}

We use 25 high-throughput sequencing data sets from several sources
to benchmark the \textbf{DeepCoDA} framework. These data come from
two collections. The first contains 13 data sets from \cite{quinn_interpretable_2020},
curated to benchmark compositional data analysis methods for microbiome
and similar data\footnote{ Available from https://zenodo.org/record/3378099/}.
The second contains 12 data sets from \cite{vangay_microbiome_2019},
curated to benchmark machine learning methods for microbiome data\footnote{ Available from https://knights-lab.github.io/MLRepo/}.
These 12 were selected from a larger collection because they each
had $\geq100$ samples, $\geq50$ samples in the smallest class, and
a binary outcome. Since log-contrasts are undefined if any feature
equals zero, we first replace zeros with a very small number (using
a method that preserves log-ratio abundance for all ratios without
zeros \cite{palarea-albaladejo_zcompositions_2015}). The data sets
have a median of 220 samples (IQR: 164-404) and 885 features (IQR:
188-1302). A thorough description of the data is available as a \textbf{Supplemental
Table}.

We develop the model in two stages. First, we use a ``discovery set''
of 13 data sets to design the architecture and choose its hyper-parameters.
Second, we use a ``verification set'' of 12 unseen data sets to
benchmark the final model. Most real data sets lack sufficient samples
to tune hyper-parameters. Hence, we use the two-step discovery/verification
procedure to identify a set of hyper-parameters that we can recommend
for real-world applications.

We benchmark all models with and without self-explanation, and report
the AUC distribution across 20 random 90\%-10\% training-test set
splits.

\subsubsection{Baselines}

For the simulated data, we train a regularized logistic regression
model separately for the absolute and relative abundances (i.e., LASSO,
with $\lambda$ chosen by cross-validation). We compare the absolute
value of the coefficients for these models to provide an intuition
for why a routine analysis of compositional data can yield spurious
interpretations.

For the real data, we benchmark our model against several baseline
log-ratio transformations, using either regularized logistic regression
or random forest. The transformations include: (a) no transformation,
(b) the centered log-ratio \cite{aitchison_statistical_1986}, (c)
principal balances \cite{pawlowsky-glahn_principal_2011}, and (d)
distal balances \cite{quinn_interpretable_2020}.

\subsection{Study 1: Feature attribution is unreliable for compositional data}

The simulated data contain absolute abundances and relative abundances.
For real data, only the relative abundances are known. Our aim is
to develop a model in which the interpretation of the relative data
agrees with the absolute data. It is easy to show that this is not
the case for even simple models like logistic regression.

\begin{figure}[th]
\begin{centering}
\includegraphics[scale=0.4]{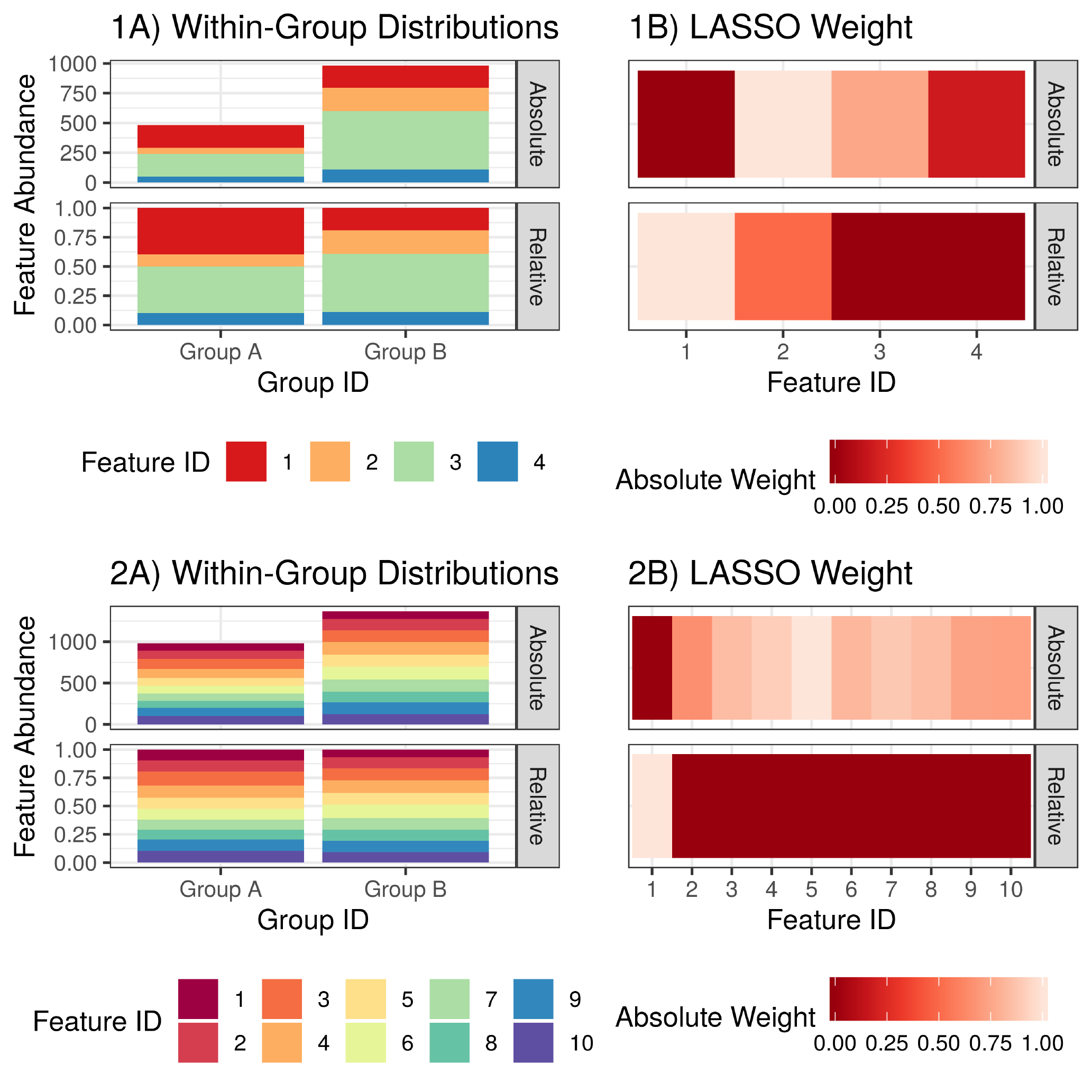}
\par\end{centering}
\caption{The top and bottom panels correspond to the two simulated data sets.
The left panels show a barplot of the absolute and relative abundances.
Importantly, both data sets have one feature that does not change
between samples (i.e., Feature ID 1), though its proportion does.
The right panels show a heatmap of the regression coefficient magnitudes
for each feature (min-max scaled, where 1 indicates a large absolute
value). Here, we see how the LASSO weights for the relative data disagree
completely with those for the absolute data. On the other hand, log-contrasts
must be the same for both the absolute and relative data, and so the
model coefficients must be the same too.}
\label{fig:simulated}
\end{figure}
Figure~\ref{fig:simulated} shows a barplot of the absolute and relative
abundances, alongside a heatmap of the regression coefficient magnitudes
for each feature. Importantly, both data sets have one feature that
does not change between samples (i.e., Feature ID 1), though its proportion
does. When fitting a regression on the absolute data, this feature
receives zero weight (as it should). However, when fitting a regression
on the relative data, this feature receives a large weight. This is
because the different groups have a different total abundance. Although
this feature has no signal, its relative abundance can perfectly differentiate
between the two groups because its relative abundance is conditional
upon all other features. Consequently, the coefficient weights swap
completely: the zeros become large and the large become zero. The
coefficients will change further when working with a subset of the
composition; this is especially a problem for microbiome data because
the sequencing assay may not measure all microorganisms (and thus
one is almost always analyzing a sub-composition).

On the other hand, log-contrasts must be the same for both the absolute
and relative data, and so the model coefficients must be the same
too. For example, consider the absolute measurements $[a=4, b=10, c=6]$,
having the closed proportions $[a=0.2, b=0.5, c=0.3]$. In both cases,
the log-contrasts (e.g., $\log(\sqrt{ab}/c)$) are identical.

\subsection{Study 2: DeepCoDA achieves state-of-the-art performance but adds
personalization}

\begin{figure}[th]
\centering{}\includegraphics[scale=0.3]{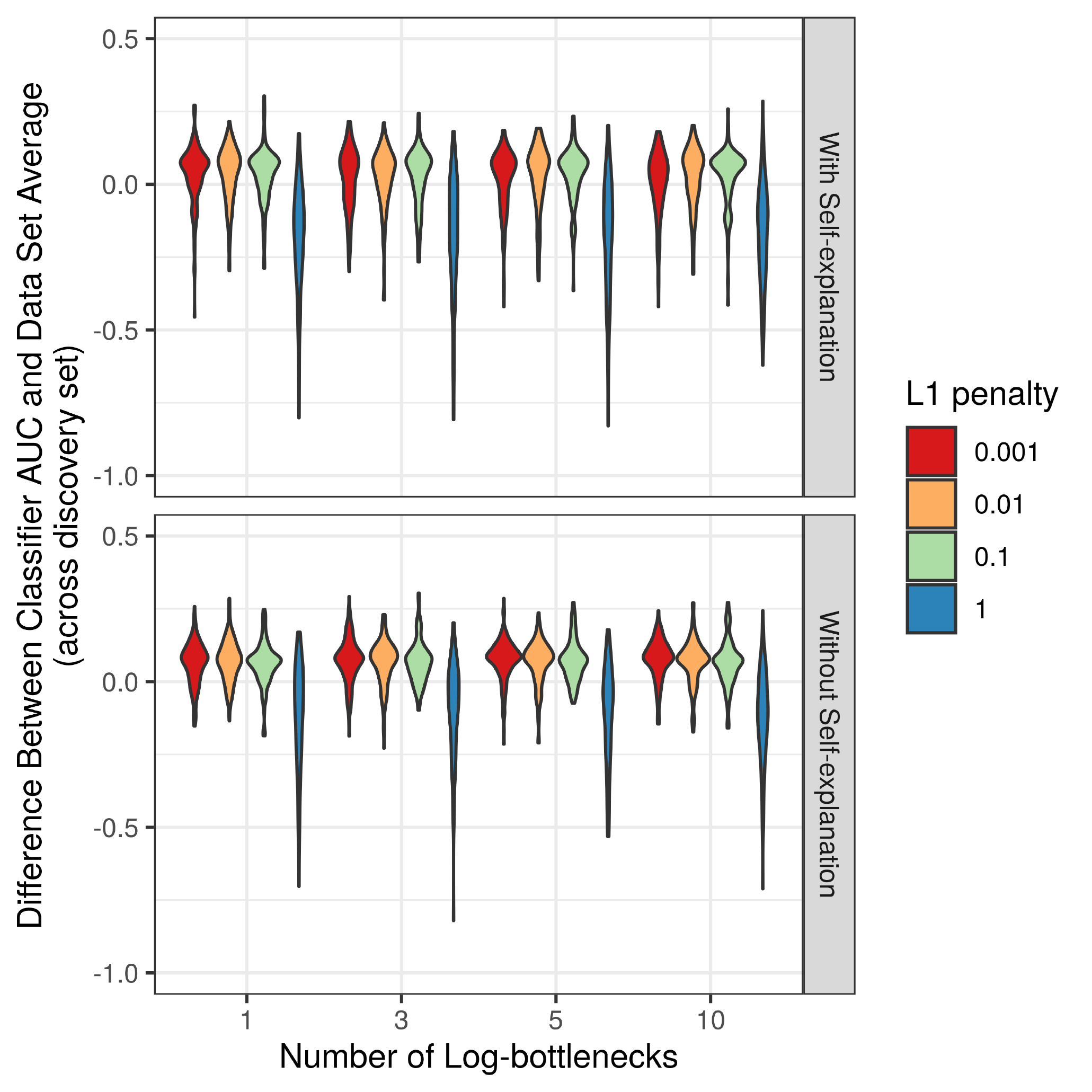} \caption{This figure shows the relative performance of our model for each combination
of hyper-parameters, with and without self-explanation (using the
``discovery set'' only). The y-axis shows the distribution of performances,
standardized so that the performances from one data set are centered
around the data set average.}
\label{fig:tuning}
\end{figure}

The data sets currently available for microbiome research typically
contain only 100s of samples, making hyper-parameter tuning infeasible.
For this reason, we sought to identify a set of hyper-parameters that
achieved good performance on a collection of data sets. Using a ``discovery
set'' of 13 data sets, we trained models with $B=[1,3,5,10]$ log-bottlenecks
and a $\lambda_{s}=[0.001,0.01,0.1,1]$ L1 penalty. Figure~\ref{fig:tuning}
shows the standardized performance for all ``discovery set'' models
for each hyper-parameter combination. Here, we see that $B=5$ and
$\lambda_{s}=0.01$ works well with or without self-explanation.
\textbf{Supplemental Figure 1} shows the \textbf{DeepCoDA }performance
compared with the baselines, where our model achieves appreciable
performance.

\begin{figure*}
\centering{}\includegraphics[scale=0.35]{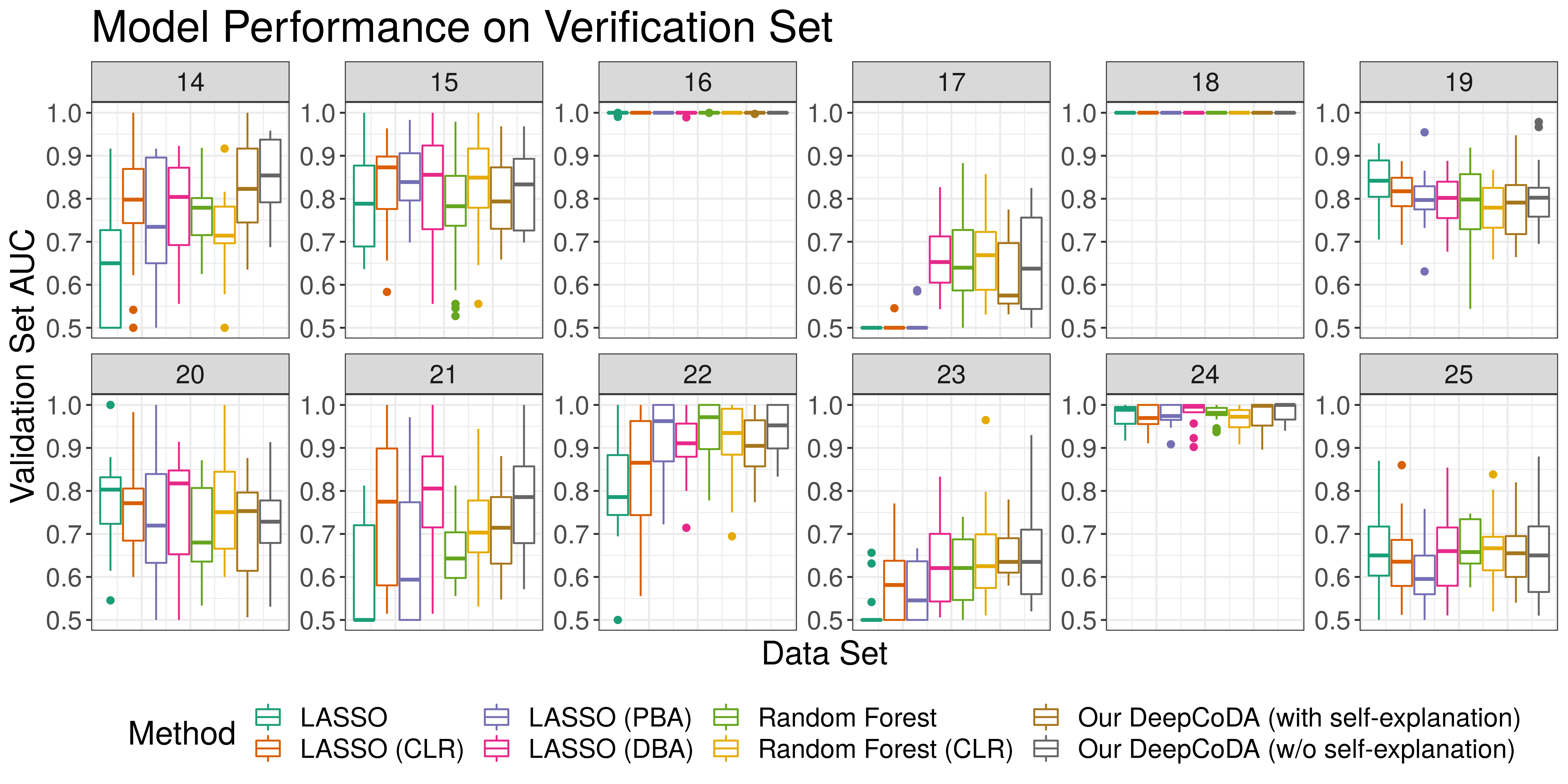} \caption{This figure shows the AUC for several models, organized by the method
(x-axis) and data set source (facet). For all models, the boxplot
shows the AUC distribution across 20 random 90\%-10\% training-test
set splits. All \textbf{DeepCoDA} models use 5 log-bottlenecks and
an L1 penalty of 0.01, chosen based on the ``discovery set''. Our
model achieves appreciable performance across the 25 data sets. See
the \textbf{Supplement} for all results from data sets 1-13.}
\label{fig:performance1}
\end{figure*}
Next, we used a ``verification set'' of 12 unseen data sets to ensure
that the selected hyper-parameters generalize to new data. Figure~\ref{fig:performance1}
shows the \textbf{DeepCoDA }performance compared with the baselines.
Again, we see that our model achieves appreciable performance. With
only 100s of samples, it is not surprising that the neural network
does not clearly outperform linear models; however, our aim is not
to improve performance, but to extend personalized interpretability
to compositional data.

Although most biomarker data sets only contain 100s of samples, \textbf{Supplemental
Figure 2} shows that \textbf{DeepCoDA }can scale to larger data sets
with 1000s of samples.

\subsection{Study 3: DeepCoDA produces transparent models}

To better understand how our model works for real data, we analyze
the network weights and layer activations for data set ``3'' (comparing
inflammatory bowel disease with healthy controls \cite{franzosa_gut_2019}).
Having already established the model's performance, we analyze the
model as trained on the entire data set of 220 samples.

\begin{figure*}[t]
\centering{}\includegraphics[scale=0.4]{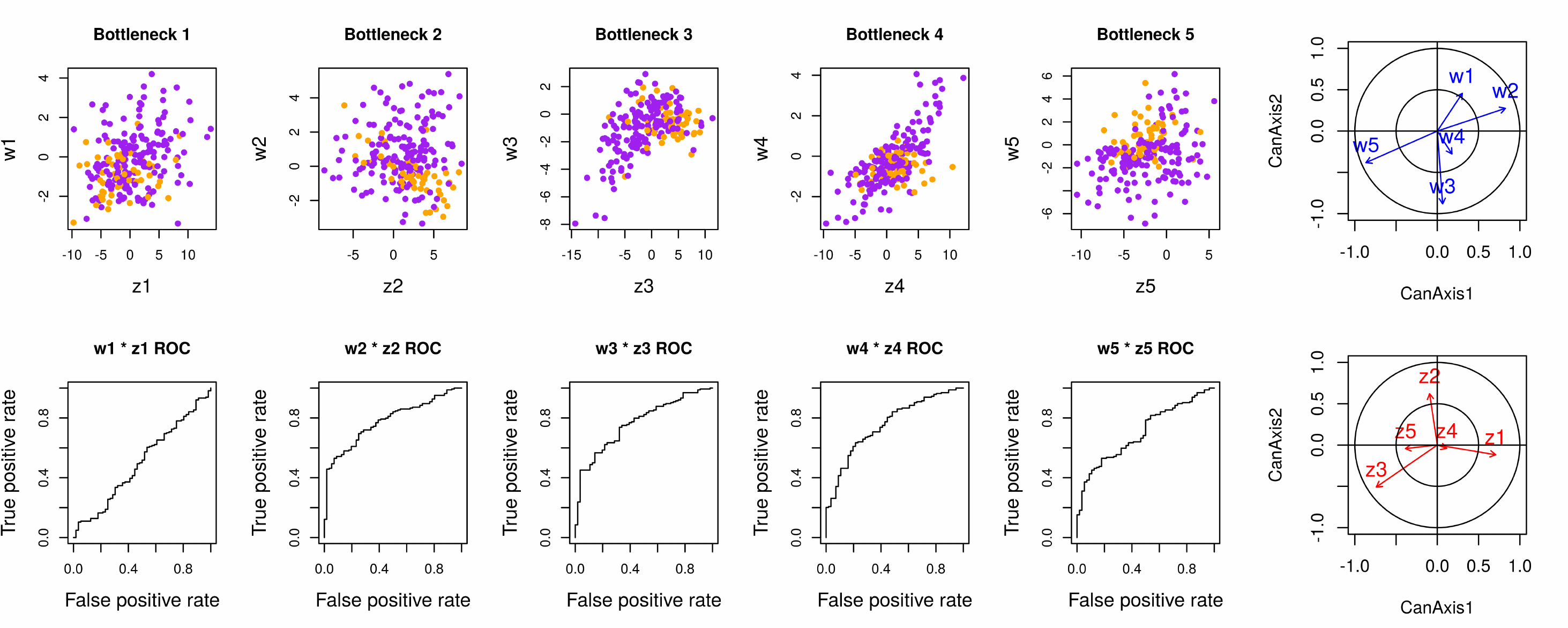}
\caption{This figure illustrates how the model makes a prediction (using data
set ``3'' as a case study). The top row has 5 panels, one for each
log-bottleneck, that plot the patient-specific weight for each log-contrast
$w_{ib}$ (y-axis) versus the value of the log-contrast itself $z_{ib}$
(x-axis). The bottom row also has 5 panels, showing the training set
ROC for the product $w_{ib}*z_{ib}$. The right-most column shows
a canonical correlation analysis between the patient-specific weights
and log-contrast values.}
\label{fig:rda}
\end{figure*}
Figure~\ref{fig:rda} illustrates how the model makes a prediction.
The top row has 5 panels, one for each log-bottleneck, that plot the
patient-specific weight for each log-contrast $w_{ib}$ (y-axis) versus
the value of the log-contrast itself $z_{ib}$ (x-axis). The bottom
row also has 5 panels, showing the training set ROC for the product
$w_{ib}*z_{ib}$. For \textbf{DeepCoDA} with self-explanation, the
final prediction is $y_i = \Phi(\sum_{b=1}^5 w_{ib}*z_{ib})$. This
formulation allows us to describe in simple algebraic terms how the
classifier made each prediction, bringing transparency to the decision-making
process.

Although the principal advantage of our model is personalized interpretability,
we can analyze the patient-specific weights to better understand how
the model ``sees'' the data. When comparing the patient-specific
weights $w_b$ with the log-contrasts $z_b$, there are (at least)
3 possibilities: (a) $w_b$ is uniform, meaning that the importance
of a log-contrast is the same for all samples, (b) $w_b$ depends
on $z_b$, meaning that the importance of a log-contrast depends on
itself (i.e., a non-linear transform of $z_b$), (c) $w_b$ depends
on other log-contrasts (i.e., a statistical interaction with $z_b$).

In Figure~\ref{fig:rda}, we see that log-contrast 4 is an example
of (b). As the ``balance'' between the numerator and denominator
``tips'' toward the numerator, the log-contrast receives a higher
positive weight. Thus, the product $w_4*z_4 \approx (z_4)^2$. On
the other hand, log-contrast 5 is an example of (c). This is shown
by the canonical correlation between the patient-specific weights
and the log-contrast values, where we see that $w_5$ is strongly
correlated with $z_3$, implying that the importance of $z_5$ depends
on $z_3$. Thus, the product $w_5 * z_5 \approx z_3 * z_5$ (because
$w_5 \approx z_3$), suggesting a log-multiplicative interaction within
the log-additive model.

Although each outcome is predicted by a linear combination of the
input, the distribution of sample weights can be correlated with the
sample input to reveal higher-order biological interactions. Therefore,
our model not only realizes personalized interpretability through
patient-specific weights, but can be studied post-hoc to reveal \emph{how}
the model generated the weights.

Our results suggest that self-explanation can produce highly interpretable
linear estimates from familiar polynomial expressions, such as power-law
transformations and statistical interactions. This finding makes a
clear connection between self-explanation and classical statistics,
and may help elicit trust from clinicians who are skeptical of deep
learning.

\subsection{Study 4: 2-levels of interpretability}

The \textbf{DeepCoDA} framework offers 2-levels of interpretability:
(1) the \textquotedbl weights\textquotedbl{} of the self-explanation
module tell us how the classifier predicts a class label; (2) the
weights of the log-bottleneck module tell us which features contribute
to each log-contrast.

At the first level, the product scores ($w_i*z_i$) can be interpreted
directly by a clinical laboratory or researcher to identify which
features drive the final prediction. When a classifier makes a decision,
the patient-specific weights ($w_i$) are multiplied with the log-contrast
values ($z_i$), then added together. In data set ``3'', a patient
is predicted to have inflammatory bowel disease if this sum exceeds
zero; otherwise, the patient is healthy. The largest product scores
contribute most to the decision.

At the second level, the log-bottleneck \emph{weights} define how
each bacteria contribute to the log-contrasts. For log-contrast 3,
the bacteria \emph{Gordonibacter pamelaeae} makes the largest contribution
to the numerator, while \emph{Bacteroides cellulosilyticus} makes
the largest contribution to the denominator. Meanwhile, the signs
of the log\emph{-}contrasts reveal which bacteria dominate. Negative
values mean that the denominator bacteria outweigh those in the numerator;
positives mean that the numerator outweighs the denominator.

We expand this discussion in the \textbf{Supplement}.
\vspace{3mm}

\section{Conclusion\label{sec:Conclusion}}

The \textbf{DeepCoDA }framework achieves personalized interpretability
through patient-specific weights, while using log-contrasts to ensure
that any interpretation is coherent for compositional data. Our model
had appreciable performance across 25 data sets when compared with
well-known baselines, and can even perform well without any hyper-parameter
tuning. However, our aim is not to improve performance, but to extend
personalized interpretability to compositional data. For \textbf{DeepCoDA},
self-explanation allows us to describe in simple algebraic terms how
the classifier made each prediction, bringing transparency to the
decision-making process.

We introduced the log-bottleneck as a new neural network architecture
to learn log-contrasts through gradient descent. However, even with
L1 regularization, the coefficients for the learned log-contrasts
were not very sparse (e.g., some log-contrasts for data set ``3''
contained 144/153 parts). Fortunately, the distribution of powers
is not uniform within a log-contrast, making it possible to summarize
a log-contrast by its most prominent members. Still, simpler log-contrasts
are preferable in most cases. Future work should examine how best
to innovate the log-bottleneck to achieve simpler log-contrasts, for
example through L0 regularization or a discrete parameter search.

\section{Availability of data and materials}

Our implementation of \textbf{DeepCoDA }is available from \url{http://github.com/nphdang/DeepCoDA}.
The scripts used to synthesize data, apply baselines, and visualize
results are available from \url{http://doi.org/10.5281/zenodo.3893986}.

\section{Acknowledgements}

This research was partially funded by the Australian Government through
the Australian Research Council (ARC)\LyXZeroWidthSpace \LyXZeroWidthSpace .
Prof Venkatesh is the recipient of an ARC Australian Laureate Fellowship
(FL170100006).

\bibliographystyle{plainnat}
\bibliography{icml2020}

\newpage{}

\appendix
\onecolumn
\title{\emph{Supplemental Material for}: \\\textbf{DeepCoDA: personalized
interpretability for \\compositional health data}}
\maketitle

\section{Study 4: 2-levels of interpretability (expanded)}

The \textbf{DeepCoDA} framework offers 2-levels of interpretability:
(1) the \textquotedbl weights\textquotedbl{} of the self-explanation
module (layer $w$ in Figure 1 of the main paper) tell us how the
classifier predicts a class label; (2) the weights of the log-bottleneck
module ($\theta_z$ in Figure 1 of the main paper) tell us which features
contribute to each log-contrast.

At the \textbf{first level}, the product scores ($w_i*z_i$) can be
interpreted directly by a clinical laboratory or researcher to identify
which features drive the final prediction. When a classifier makes
a decision, the patient-specific weights ($w_i$) are multiplied with
the log-contrast values ($z_i$), then added together. In data set
``3'', a patient is predicted to have inflammatory bowel disease
if this sum exceeds zero; otherwise, the patient is healthy. The largest
product scores contribute most to the decision.

Consider two patients, chosen randomly:

\begin{itemize}
\item Patient 26 is healthy: their product scores are [2.0, -15.6, -3.6, -1.1, 1.8]. 
\item Patient 13 is unhealthy: their product scores are [0.5, -7.3, 8.8, 7.4, 0.1].
\end{itemize}

For patient 26, the second term is highly negative, suggesting that
the patient is healthy. Since the second term is derived from the
second log-contrast, we can infer that log-contrast 2 is most important
for this prediction.

For patient 13, the third and fourth terms are highly positive, suggesting
that the patient is unhealthy. Interestingly, the second is highly
negative (like patient 26), suggesting that the second log-contrast
\textquotedblleft looks\textquotedblright{} healthy. However, this
negative score is not enough to sway the final decision.

At the \textbf{second level}, the log-bottleneck weights define how
each bacteria contribute to the ratios. For log-contrast 3, the bacteria
\emph{Gordonibacter pamelaeae} makes the largest contribution to the
numerator, while \emph{Bacteroides cellulosilyticus} makes the largest
contribution to the denominator. For patient 13, the log-contrast
values are {[}0.61, -2.98, -4.64, 5.79, -1.76{]}. The signs of the
log\emph{-}contrasts reveal which bacteria dominate. Negative values
mean that the denominator bacteria outweigh those in the numerator;
positives mean that the numerator outweighs the denominator.

Meanwhile, our canonical correlation analysis reminds us that the
importance of log-contrast 3 ($w_3$) depends on the value of log-contrast
2 ($z_2$), via an interaction learned automatically from the data.
The top panels in Figure 5 summarize the distribution of these patient-specific
weights and log-contrast values across the entire patient cohort.

\begin{figure}
\centering{}\includegraphics[scale=0.45]{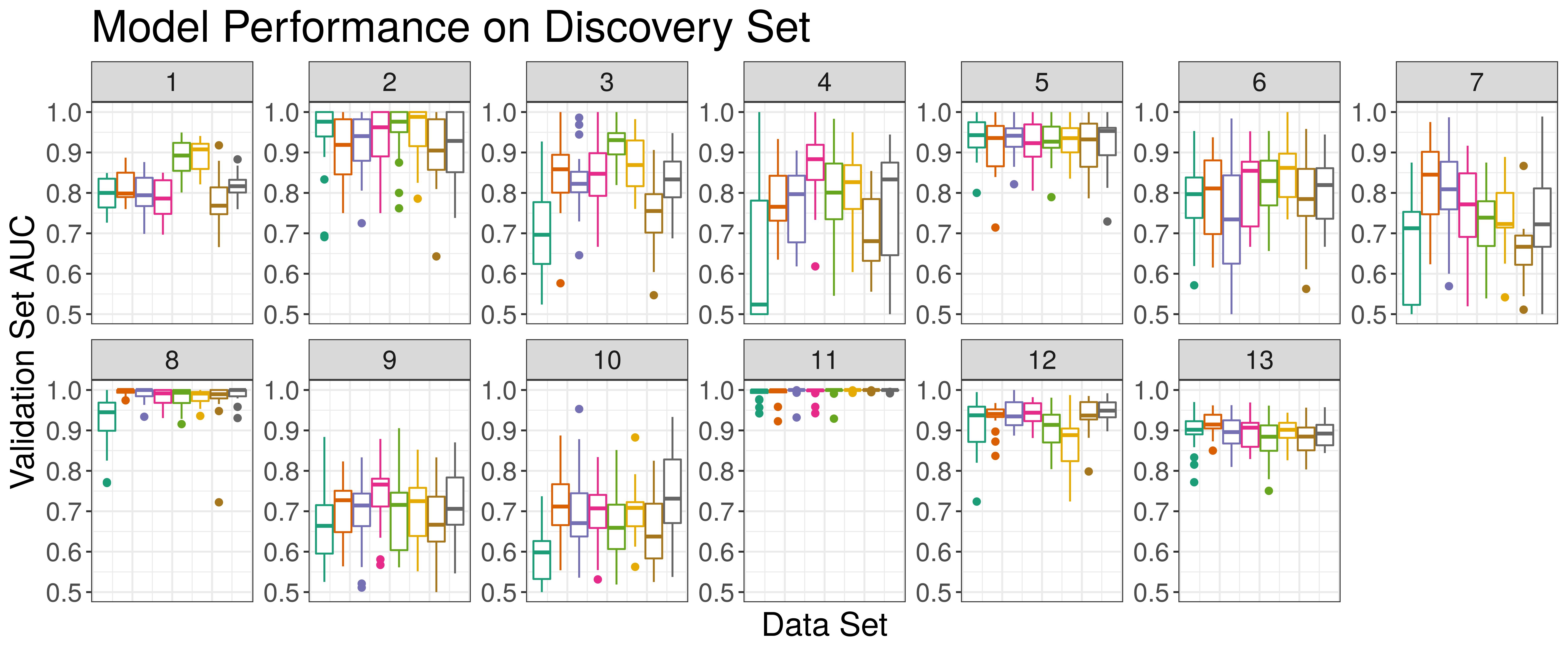} \caption{This figure shows the AUC for several models, organized by the method
(x-axis) and data set source (facet). For all models, the boxplot
shows the AUC distribution across 20 random 90\%-10\% training-test
set splits. All \textbf{DeepCoDA} models use 5 log-bottlenecks and
an L1 penalty of 0.01, chosen based on the ``discovery set''. Our
model achieves appreciable performance across the 25 data sets. However,
our aim is not to improve performance, but to extend personalized
interpretability to compositional data.}
\end{figure}
\begin{figure}
\centering{}\includegraphics[scale=0.45]{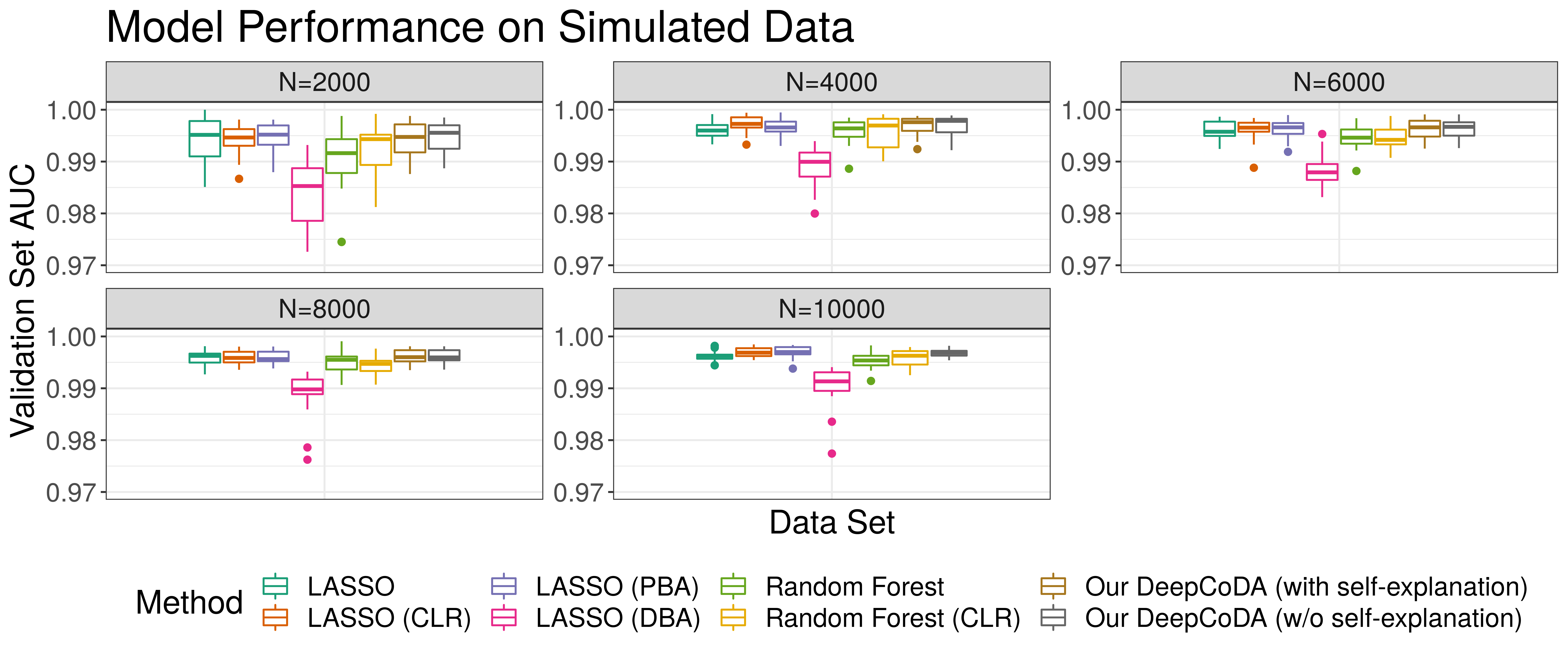} \caption{This figure shows the AUC for several models, organized by the method
(x-axis) and number of samples in the second synthetic data set (facet).
For all models, the boxplot shows the AUC distribution across 20 random
90\%-10\% training-test set splits. This figure confirms that the
\textbf{DeepCoDA} model can scale to larger data sets with many samples.}
\end{figure}
\begin{table}
\begin{centering}
\begin{tabular}{|r|r|r|r|l|l|}
\hline 
\textbf{Data set ID} & \textbf{\# Samples} & \textbf{\# Features} & \textbf{\# Classes} & \textbf{Class 1} & \textbf{Class 2}\tabularnewline
\hline 
\hline 
1 & 975 & 48 & 2 & Crohn\textquoteright s disease & Without\tabularnewline
\hline 
2 & 128 & 60 & 2 & Men who have sex with men & Without\tabularnewline
\hline 
3 & 220 & 153 & 2 & Control & IBD\tabularnewline
\hline 
4 & 164 & 158 & 2 & Crohn\textquoteright s disease & Ulcerative colitis\tabularnewline
\hline 
5 & 220 & 885 & 2 & Control & IBD\tabularnewline
\hline 
6 & 164 & 885 & 2 & Crohn\textquoteright s disease & Ulcerative colitis\tabularnewline
\hline 
7 & 182 & 278 & 2 & Case & Diarrheal control\tabularnewline
\hline 
8 & 247 & 610 & 2 & Case & Non-Diarrheal control\tabularnewline
\hline 
9 & 292 & 1133 & 2 & Colorectal cancer (CRC) & Without\tabularnewline
\hline 
10 & 318 & 1302 & 2 & Colorectal cancer (CRC) & Non-CRC control\tabularnewline
\hline 
11 & 1182 & 188 & 2 & Primary solid tumor & Solid tissue normal\tabularnewline
\hline 
12 & 1004 & 188 & 2 & Her2 Cancer & Not Her2 Cancer\tabularnewline
\hline 
13 & 718 & 188 & 2 & LumA Cancer & LumB Cancer\tabularnewline
\hline 
14 & 140 & 992 & 2 & Crohn\textquoteright s disease (ileum) & Without (ileum)\tabularnewline
\hline 
15 & 160 & 992 & 2 & Crohn\textquoteright s disease (rectum) & Without (rectum)\tabularnewline
\hline 
16 & 2070 & 3090 & 2 & GI tract & Oral\tabularnewline
\hline 
17 & 180 & 3090 & 2 & Female & Male\tabularnewline
\hline 
18 & 404 & 3090 & 2 & Stool & Tongue (dorsum)\tabularnewline
\hline 
19 & 408 & 3090 & 2 & Subgingival plaque & Supragingival plaque\tabularnewline
\hline 
20 & 172 & 980 & 2 & Healthy & Colorectal cancer\tabularnewline
\hline 
21 & 124 & 2526 & 2 & Without & Diabetes\tabularnewline
\hline 
22 & 130 & 2579 & 2 & Cirrhosis & Without\tabularnewline
\hline 
23 & 199 & 660 & 2 & Black & Hispanic\tabularnewline
\hline 
24 & 342 & 660 & 2 & Nugent score high & Nugent score low\tabularnewline
\hline 
25 & 200 & 660 & 2 & Black & White\tabularnewline
\hline 
\end{tabular}
\par\end{centering}
\caption{Characteristics of the 25 data sets used in Section 4.3.}
\end{table}
\begin{table}
\centering{}%
\begin{tabular}{|c|l|l|}
\hline 
\textbf{Data set ID} & \textbf{Original source} & \textbf{Retrieved via}\tabularnewline
\hline 
\hline 
1 & doi: 10.1016/j.chom.2014.02.005 & doi: 10.1128/mSystems.00053-18\tabularnewline
\hline 
2 & doi: 10.1016/j.ebiom.2016.01.032 & doi: 10.1128/mSystems.00053-18\tabularnewline
\hline 
3 & doi: 10.1038/s41564-018-0306-4 & supplemental materials\tabularnewline
\hline 
4 & doi: 10.1038/s41564-018-0306-4 & supplemental materials\tabularnewline
\hline 
5 & doi: 10.1038/s41564-018-0306-4 & supplemental materials\tabularnewline
\hline 
6 & doi: 10.1038/s41564-018-0306-4 & supplemental materials\tabularnewline
\hline 
7 & doi: 10.1128/mBio.01021-14 & doi: doi.org/10.1038/s41467-017-01973-8\tabularnewline
\hline 
8 & doi: 10.1128/mBio.01021-15 & doi: doi.org/10.1038/s41467-017-01973-9\tabularnewline
\hline 
9 & doi: 10.1186/s13073-016-0290-3 & doi: doi.org/10.1038/s41467-017-01973-10\tabularnewline
\hline 
10 & doi: 10.1186/s13073-016-0290-3 & doi: doi.org/10.1038/s41467-017-01973-11\tabularnewline
\hline 
11 & doi: 10.1038/ng.2764 & labels from doi: 10.1186/s13058-016-0724-2\tabularnewline
\hline 
12 & doi: 10.1038/ng.2764 & labels from doi: 10.1186/s13058-016-0724-2\tabularnewline
\hline 
13 & doi: 10.1038/ng.2764 & labels from doi: 10.1186/s13058-016-0724-2\tabularnewline
\hline 
14 & doi: 10.1016/j.chom.2014.02.005 & doi: 10.1093/gigascience/giz042\tabularnewline
\hline 
15 & doi: 10.1016/j.chom.2014.02.005 & doi: 10.1093/gigascience/giz043\tabularnewline
\hline 
16 & doi: 10.1038/nature11209 & doi: 10.1093/gigascience/giz044\tabularnewline
\hline 
17 & doi: 10.1038/nature11209 & doi: 10.1093/gigascience/giz045\tabularnewline
\hline 
18 & doi: 10.1038/nature11209 & doi: 10.1093/gigascience/giz046\tabularnewline
\hline 
19 & doi: 10.1038/nature11209 & doi: 10.1093/gigascience/giz047\tabularnewline
\hline 
20 & doi: 10.1101/gr.126573.111 & doi: 10.1093/gigascience/giz048\tabularnewline
\hline 
21 & doi: 10.1038/nature11450 & doi: 10.1093/gigascience/giz049\tabularnewline
\hline 
22 & doi: 10.1038/nature13568 & doi: 10.1093/gigascience/giz050\tabularnewline
\hline 
23 & doi: 10.1073/pnas.1002611107 & doi: 10.1093/gigascience/giz051\tabularnewline
\hline 
24 & doi: 10.1073/pnas.1002611107 & doi: 10.1093/gigascience/giz052\tabularnewline
\hline 
25 & doi: 10.1073/pnas.1002611107 & doi: 10.1093/gigascience/giz053\tabularnewline
\hline 
\end{tabular}\caption{Sources for the 25 data sets used in Section 4.3.}
\end{table}

\end{document}